\documentclass[11pt,a4paper]{article}
\usepackage[hyperref]{acl2018}
\usepackage{times}
\usepackage{latexsym}
\usepackage{graphicx}
\usepackage{url}
\usepackage{bm}
\usepackage{amsmath}
\usepackage{multirow}
\usepackage{subfig}
\usepackage{mathtools}
\usepackage{float}
\usepackage{enumitem}

\aclfinalcopy 
\def\aclpaperid{961}

\title{Neural Argument Generation \\Augmented with Externally Retrieved Evidence}

\author{Xinyu Hua \and Lu Wang \\ College of Computer and Information Science \\ Northeastern University \\ Boston, MA 02115 \\ {\tt hua.x@husky.neu.edu} \quad {\tt luwang@ccs.neu.edu}}

\begin{document}
\maketitle

\begin{abstract}
\fontsize{10}{12}\selectfont
High quality arguments are essential elements for human reasoning and decision-making processes. However, effective argument construction is a challenging task for both human and machines. In this work, we study a novel task on {\it automatically generating arguments of a different stance for a given statement}. 
We propose an encoder-decoder style neural network-based argument generation model enriched with externally retrieved evidence from Wikipedia. 
Our model first generates a set of talking point phrases as intermediate representation, followed by a separate decoder producing the final argument based on both input and the keyphrases. 
Experiments on a large-scale dataset collected from Reddit show that our model constructs arguments with more topic-relevant content than a popular sequence-to-sequence generation model according to both automatic evaluation and human assessments. 
\end{abstract}

\section{Introduction}
Generating high quality arguments plays a crucial role in decision-making and reasoning processes~\cite{bonet1996arguing,byrnes2013nature}. A multitude of arguments and counter-arguments are constructed on a daily basis, both online and offline, to persuade and inform us on a wide range of issues. For instance, debates are often conducted in legislative bodies to secure enough votes for bills to pass. In another example, online deliberation has become a popular way of soliciting public opinions on new policies' pros and cons~\cite{albrecht2006whose,park2012facilitative}. 
Nonetheless, constructing persuasive arguments is a daunting task, for both human and computers.
We believe that developing effective argument generation models will enable a broad range of compelling applications, including debate coaching, improving students' essay writing skills, and providing context of controversial issues from different perspectives. As a consequence, there exists a pressing need for automating the argument construction process. 

\begin{figure}[t]
    \hspace{-2mm}
    \includegraphics[width=79mm]{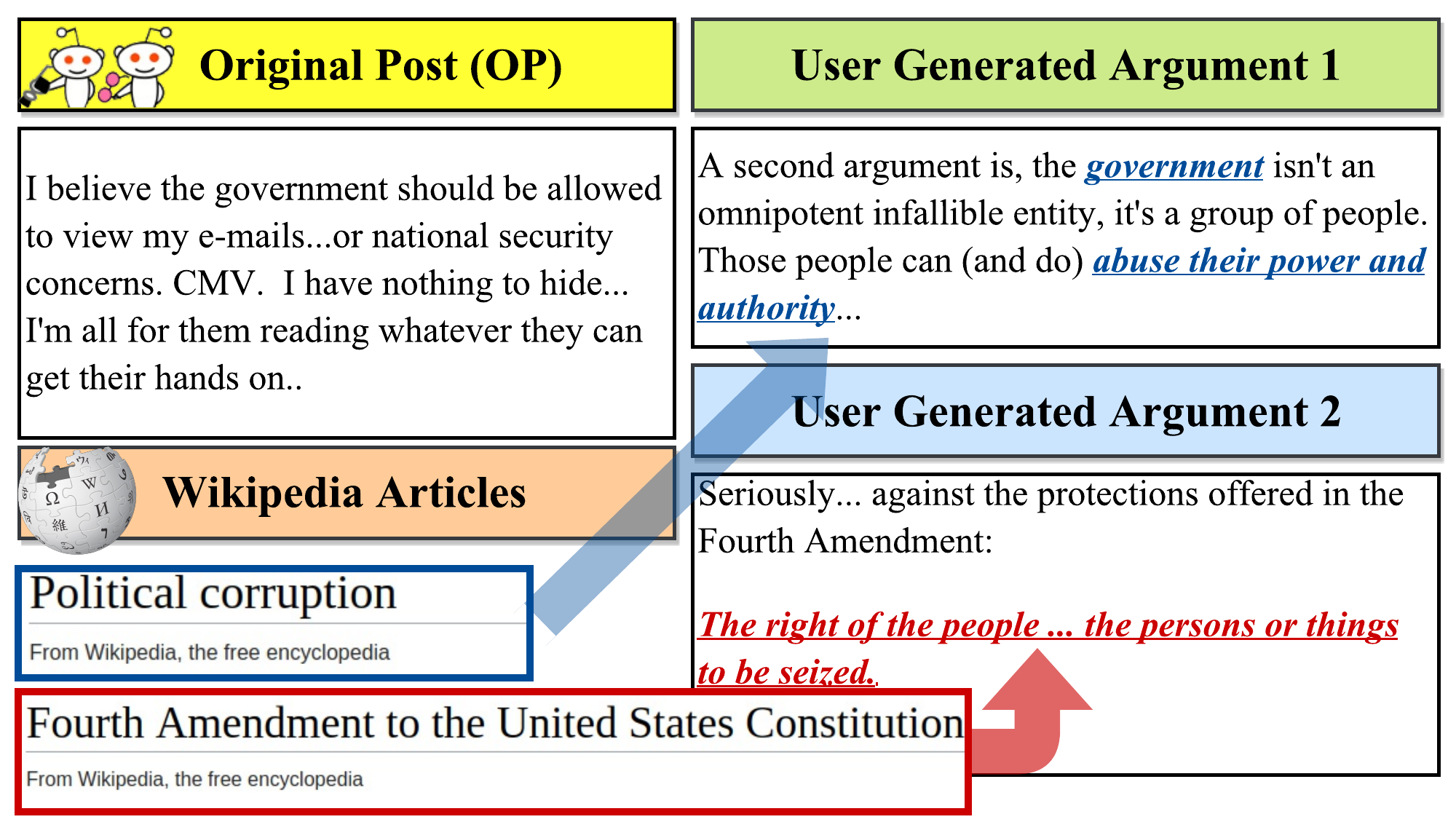}
    \caption{\fontsize{10}{12}\selectfont Sample user arguments from Reddit Change My View subcommunity that argue against original post's thesis on ``government should be allowed to view private emails''. Both arguments leverage supporting information from Wikipedia articles.
    }
    \label{fig:motivating_example}
\end{figure}

To date, progress made in argument generation has been limited to retrieval-based methods---arguments are ranked based on relevance to a given topic, then the top ones are selected for inclusion in the output~\cite{rinott-EtAl:2015:EMNLP,wachsmuth-EtAl:2017:ArgumentMining,hua-wang:2017:Short}. 
Although sentence ordering algorithms are developed for information structuring~\cite{sato-EtAl:2015:ACL-IJCNLP-2015-System-Demonstrations,reisert-EtAl:2015:ARG-MINING}, existing methods lack the ability of synthesizing information from different resources, leading to redundancy and incoherence in the output.

In general, the task of argument generation presents numerous challenges, ranging from aggregating supporting evidence to generating text with coherent logical structure. 
One particular hurdle comes from the underlying natural language generation (NLG) stack, whose success has been limited to a small set of domains. Especially, most previous NLG systems rely on templates that are either constructed by rules~\cite{hovy1993automated,belz2008automatic,bouayadagha-casamayor-wanner:2011:ENLG}, or acquired from a domain-specific corpus~\cite{angeli-liang-klein:2010:EMNLP} to enhance grammaticality and coherence. This makes them unwieldy to be adapted for new domains.

\begin{figure*}[t]
\centering
  \includegraphics[width=150mm,height=66mm]{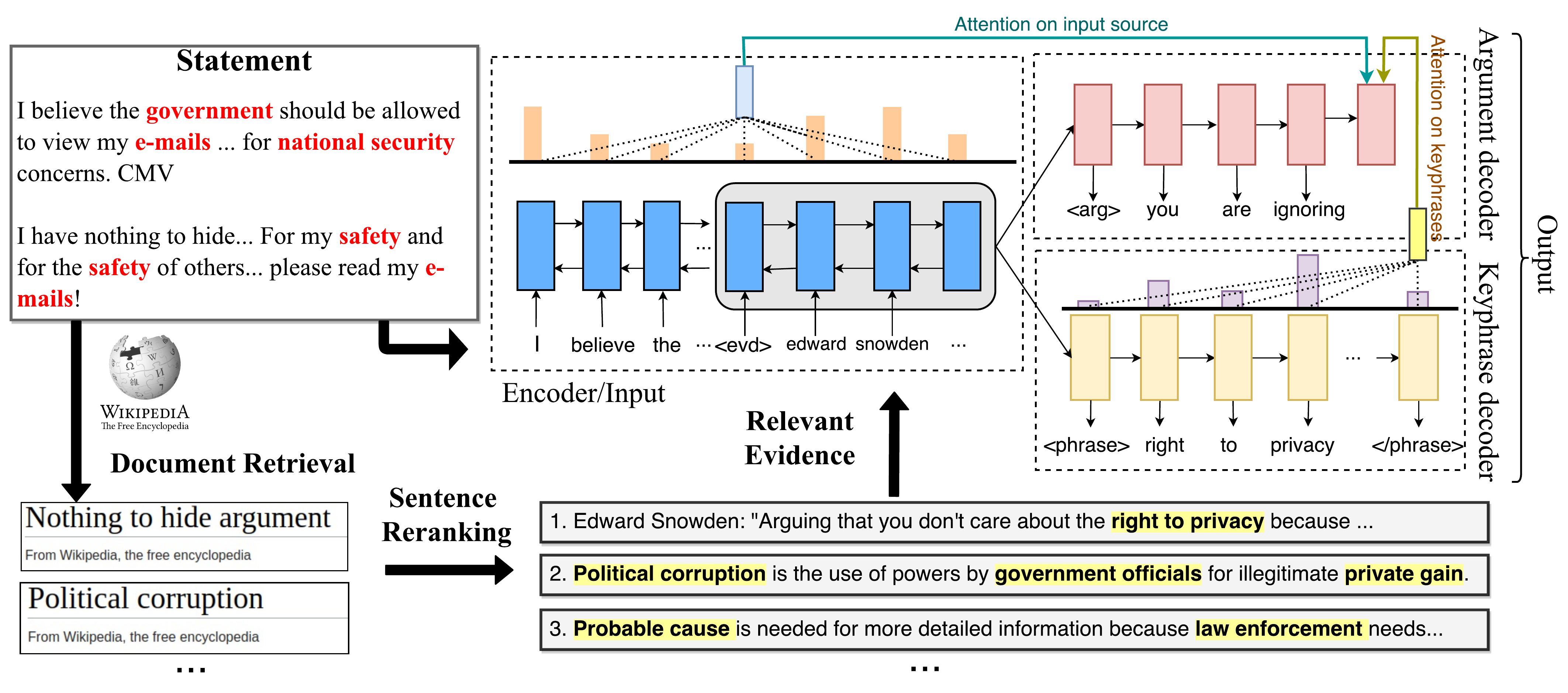}
  \caption{\fontsize{10}{12}\selectfont 
  Overview of our system pipeline (best viewed in color). Given a statement, relevant articles are retrieved from Wikipedia with topic signatures from statement as queries (marked in red and boldface). A reranking module then outputs top sentences as evidence. The statement and the evidence (encoder states in gray panel) are concatenated and encoded as input for our argument generation model. During decoding, the keyphrase decoder first generates talking points as phrases, followed by the argument decoder which constructs the argument by attending both input and keyphrases. 
  }
   \label{fig:pipeline}
\end{figure*}

In this work, we study the following novel problem: {\it given a statement on a controversial issue, generate an argument of an alternative stance}. 
To address the above challenges, we present {\it a neural network-based argument generation framework augmented with externally retrieved evidence}. Our model is inspired by the observation that when humans construct arguments, they often collect
references from external sources, e.g., Wikipedia or research papers, and then write their own arguments by synthesizing talking points from the references. 
Figure~\ref{fig:motivating_example} displays sample arguments by users from Reddit subcommunity \texttt{/r/ChangeMyView} 
\footnote{ \url{https://www.reddit.com/r/changemyview}} 
who argue against the motion that ``government should be allowed to view private emails''. Both replies leverage information drawn from Wikipedia, such as ``political corruption'' and ``Fourth Amendment on protections of personal privacy". 

Concretely, our neural argument generation model adopts the popular encoder-decoder-based sequence-to-sequence (seq2seq) framework~\cite{sutskever2014sequence}, which has achieved significant success in various text generation tasks~\cite{bahdanau2014neural,wen-EtAl:2015:EMNLP,wang-ling:2016:N16-1,mei-bansal-walter:2016:N16-1,wiseman-shieber-rush:2017:EMNLP2017}. Our encoder takes as input a statement on a disputed issue, and a set of relevant evidence automatically retrieved from English Wikipedia\footnote{ \url{https://en.wikipedia.org/}}. Our decoder consists of two separate parts, one of which first generates keyphrases as intermediate representation of ``talking points'', and the other then generates an argument based on both input and keyphrases. 

Automatic evaluation based on BLEU~\cite{papineni-EtAl:2002:ACL} shows that our framework generates better arguments than directly using retrieved sentences or popular seq2seq-based generation models~\cite{bahdanau2014neural} that are also trained with retrieved evidence. 
We further design a novel evaluation procedure to measure whether the arguments are on-topic by predicting their relevance to the given statement based on a separately trained relevance estimation model. Results suggest that our model generated arguments are more likely to be predicted as on-topic, compared to other seq2seq-based generations models.

The rest of this paper is organized as follows. Section \ref{sec:framework} highlights the roadmap of our system. The dataset used for our study is introduced in Section \ref{sec:data}. 
The model formulation and retrieval methods are detailed in Sections \ref{sec:model} and \ref{sec:retrieval}. 
We then describe the experimental setup and results in Sections \ref{sec:experiments} and \ref{sec:results}, followed by further analysis and future directions in Section \ref{sec:discussion}. Related work is discussed in Section \ref{sec:related}. Finally, we conclude in Section \ref{sec:conclusion}.

\section{Framework}
\label{sec:framework}
Our argument generation pipeline, consisting of \emph{evidence retrieval} and \emph{argument construction}, is depicted in Figure \ref{fig:pipeline}. 
Given a statement, a set of queries are constructed based on its topic signature words (e.g., ``government'' and ``national security'') to retrieve a list of relevant articles from Wikipedia. A reranking component further extracts sentences that may contain supporting evidence, which are used as additional input information for the neural argument generation model. 

The generation model then encodes the statement and the evidence with a shared encoder in sequence. Two decoders are designed: the {\it keyphrase decoder} first generates an intermediate representation of talking points in the form of keyphrases (e.g., ``right to privacy'', ``political corruption"), followed by a separate {\it argument decoder} which produces the final argument.

\section{Data Collection and Processing}
\label{sec:data}
We draw data from Reddit subcommunity \texttt{/r/ChangeMyView} (henceforth CMV), which focuses on facilitating open discussions on a wide range of disputed issues. 
Specifically, CMV is structured as discussion threads, where the original post (OP) starts with a viewpoint on a controversial topic, followed with detailed reasons, then other users reply with counter-arguments.
Importantly, when a user believes his view has been changed by an argument, a \textit{delta} is often awarded to the reply. 

In total, 26,761 threads from CMV are downloaded, dating from January 2013 to June 2017\footnote{Dataset used in this paper is available at \url{http://xinyuhua.github.io/Resources/}. }. Only root replies (i.e., replies directly addressing OP) that meet all of the following requirements are included: (1) longer than 5 words, (2) without offensive language\footnote{ We use offensive words collected by Google's What Do You Love project: \url{https://gist.github.com/jamiew/1112488}, last accessed on February 22nd, 2018.}, (3) awarded with \textit{delta} or with more upvotes than downvotes, and (4) not generated by system moderators. 

After filtering, the resultant dataset contains 26,525 OPs along with 305,475 relatively high quality root replies. 
We treat each OP as the input statement, and the corresponding root replies as target arguments, on which our model is trained and evaluated.

\smallskip
\noindent \textbf{A Focused Domain Dataset.}  
The current dataset contains diverse domains with unbalanced numbers of arguments. 
We therefore choose samples from the politics domain due to its large volume of discussions and good coverage of popular arguments in the domain. 

However, topic labels are not available for the discussions. We thus construct a domain classifier for politics vs. non-politics posts based on a logistic regression model with unigram features, trained from our heuristically labeled Wikipedia abstracts\footnote{About $1.3$ million English 
Wikipedia abstracts are downloaded from \url{http://dbpedia.org/page/}.}. 
Concretely, we manually collect two lists of keywords that are indicative of politics and non-politics. Each abstract is labeled as politics or non-politics if its title only matches keywords from one category.\footnote{Sample keywords for politics: ``congress'', ``election'', ``constitution''; for non-politics: ``art'', ``fashion'',``music''. Full lists are provided in the supplementary material.} 
In total, 264,670 politics abstracts and 827,437 of non-politics are labeled. Starting from this dataset, our domain classifier is trained in a bootstrapping manner by gradually adding OPs predicted as politics or non-politics.\footnote{More details about our domain classifier are provided in the supplementary material.} 
Finally, 12,549 OPs are labeled as politics, each of which is paired with 9.4 high-quality target arguments on average. The average length for OPs is 16.1 sentences of 356.4 words, and 7.7 sentences of 161.1 words for arguments.

\section{Model}
\label{sec:model}
In this section, we present our argument generation model, which jointly learns to generate talking points in the form of keyphrases and produce arguments based on the input and keyphrases. Extended from the successful seq2seq attentional model~\cite{bahdanau2014neural}, our proposed model is novel in the following ways. First, two separate decoders are designed, one for generating keyphrases, the other for argument construction. By sharing the encoder with keyphrase generation, our argument decoder is better aware of salient talking points in the input. Second, a novel attention mechanism is designed for argument decoding by attending both input and the previously generated keyphrases. Finally, a reranking-based beam search decoder is introduced to promote topic-relevant generations.

\subsection{Model Formulation}
Our model takes as input a sequence of tokens $\bm{x} = \{\bm{x}^O; \bm{x}^E\}$, where $\bm{x}^O$ is the statement sequence and $\bm{x}^E$ contains relevant evidence that is extracted from Wikipedia based on a separate retrieval module. A special token \texttt{<evd>} is inserted between $\bm{x}^O$ and $\bm{x}^E$. Our model then first generates a set of keyphrases as a sequence $\bm{y}^p = \{y^p_l\}$, followed by an argument $\bm{y}^a = \{y^a_t\}$, by maximizing $\log P(\bm{y}|\bm{x})$, where $\bm{y}=\{\bm{y}^p; \bm{y}^a\}$.

The objective is further decomposed into $\sum_{t} \log P(y_t|y_{1:t-1},\bm{x})$, with each term estimated by a softmax function over a non-linear transformation of decoder hidden states $\bm{s}^a_t$ and $\bm{s}^p_t$, for argument decoder and keyphrase decoder, respectively. The hidden states are computed as done in \newcite{bahdanau2014neural} with attention: 

\vspace{-2mm}
{\fontsize{10}{11}\selectfont
\setlength{\abovedisplayskip}{2pt}
\setlength{\belowdisplayskip}{2pt}
\begin{align}
\fontsize{10}{11}\selectfont
 	 & \bm{s}_t = g(\bm{s}_{t-1}, \bm{c}_t, y_t) \label{eq:attn_0}\\
     &\bm{c}_t = \sum_{j=1}^T \alpha_{tj} \bm{h}_j \label{eq:attn_1} \\
     &\alpha_{tj} = \frac{\textnormal{exp}(e_{tj})}{\sum_{k=1}^T \textnormal{exp}(e_{tk})} \label{eq:attn_2} \\
     &e_{tj} = \bm{v}^T\tanh (\bm{W_h} \bm{h}_j + \bm{W_s} \bm{s}_t + \bm{b}_{attn}) \label{eq:attn_3}
\end{align}
}

Notice that two sets of parameters and different state update functions $g(\cdot)$ are learned for separate decoders: $\{\bm{W}^a_h$, $\bm{W}^a_s$, $\bm{b}^a_{attn}, g^a(\cdot) \}$ for the argument decoder; $\{\bm{W}^p_h$,  $\bm{W}^p_s$,  $\bm{b}^p_{attn}, g^p(\cdot) \}$ for the keyphrase decoder. 

\smallskip
\noindent \textbf{Encoder.} 
A two-layer bidirectional LSTM (bi-LSTM) is used to obtain the encoder hidden states $\bm{h}_i$ for each time step $i$. For biLSTM, the hidden state is the concatenation of forward and backward hidden states: $\bm{h}_i=[\overrightarrow{\bm{ h}_i}; \overleftarrow{\bm{h}_i}]$. Word representations are initialized with 200-dimensional pre-trained GloVe embeddings~\cite{pennington-socher-manning:2014:EMNLP2014}, and updated during training. 
The last hidden state of encoder is used to initialize both decoders. In our model the encoder is shared by argument and keyphrase decoders.

\smallskip
\noindent \textbf{Decoders.}
Our model is equipped with two decoders: \textit{keyphrase decoder} and \textit{argument decoder}, each is implemented with a separate two-layer unidirectional LSTM, in a similar spirit with one-to-many multi-task sequence-to-sequence learning~\cite{luong2015multi}. The distinction is that our training objective is the sum of two loss functions:

\vspace{-3mm}
{\fontsize{10}{11}\selectfont
\begin{align}
\begin{split}
  \mathcal{L}(\theta)  = &-\frac{\alpha}{T_p}{\sum_{(\bm{x},\bm{y}^p)\in D}} \log P(\bm{y}^p|\bm{x};\theta) \\ 
  &- \frac{(1-\alpha)}{T_a}{\sum_{(\bm{x},\bm{y}^a)\in D}} \log P(\bm{y}^{a}|\bm{x};\theta)  \label{eq:loss_overall} \\
\end{split}
\end{align}
}

\noindent where $T_p$ and $T_a$ denote the lengths of reference keyphrase sequence and argument sequence. 
$\alpha$ is a weighting parameter, and it is set as $0.5$ in our experiments.

\smallskip
\noindent \textbf{Attention over Both Input and Keyphrases.} 
Intuitively, the argument decoder should consider the generated keyphrases as talking points during the generation process. We therefore propose an attention mechanism that can attend both encoder hidden states and the keyphrase decoder hidden states. Additional context vector $\bm{c}'_t$ is then computed over keyphrase decoder hidden states $\bm{s}^p_j$, which is used for computing the new argument decoder state:

\vspace{-2mm}
{\fontsize{10}{11}\selectfont
\setlength{\abovedisplayskip}{2pt}
\setlength{\belowdisplayskip}{2pt}
\begin{align}
&\bm{s}^a_t = g'(\bm{s}^a_{t-1}, [\bm{c}_t; \bm{c}'_t], {y}^a_t) \label{eq:kattn_1} \\
& \bm{c}'_t = \sum_{j=1}^{T_p} \alpha'_{tj}\bm{s}^p_j \label{eq:kattn_3}\\
& \alpha'_{tj} = \frac{\textnormal{exp}(e'_{tj})}{\sum_{k=1}^{T_p} \textnormal{exp}{(e'_{tk})}} \label{eq:kattn_4}\\
&  e'_{tj} = {\bm{v}'}^T\tanh (\bm{W'_p} \bm{s}^p_j + \bm{W'_a} \bm{s}^a_t + \bm{b}'_{attn})  \label{eq:kattn_5}
\end{align}
}

\noindent where $\bm{s}^p_j$ is the hidden state of keyphrase decoder at position $j$, $\bm{s}^a_t$ is the hidden state of argument decoder at timestep $t$, and $\bm{c}_t$ is computed in Eq.~\ref{eq:attn_1}.

\smallskip
\noindent \textbf{Decoder Sharing.} 
We also experiment with a shared decoder between keyphrase generation and argument generation: the last hidden state of the keyphrase decoder is used as the initial hidden state for the argument decoder. A special token \texttt{<arg>} is inserted between the two sequences, indicating the start of argument generation. 

\subsection{Hybrid Beam Search Decoding} 
Here we describe our decoding strategy on the argument decoder. We design a hybrid beam expansion method combined with segment-based reranking to promote diversity of beams and informativeness of the generated arguments. 

\smallskip
\noindent \textbf{Hybrid Beam Expansion.} 
In the standard beam search, the top $k$ words of highest probability are selected deterministically based on the softmax output to expand each hypothesis. However, this may lead to suboptimal output for text generation~\cite{wiseman-rush:2016:EMNLP2016}, e.g., one beam often dominates and thus inhibits hypothesis diversity. 
Here we only pick the top $n$ words ($n<k$), and randomly draw another $k-n$ words based on the multinomial distribution after removing the $n$ expanded words from the candidates. This leads to a more diverse set of hypotheses.

\smallskip
\noindent \textbf{Segment-based Reranking.} 
We also propose to rerank the beams every $p$ steps based on beam's coverage of content words from input. Based on our observation that likelihood-based reranking often leads to overly generic arguments (e.g., ``I don't agree with you''), this operation has the potential of encouraging more informative generation. 
$k=10$, $n=3$, and $p=10$ are used for experiments. The effect of parameter selection is studied in Section~\ref{sec:results}.

\section{Relevant Evidence Retrieval}
\label{sec:retrieval}
\subsection{Retrieval Methodology}
We take a two-step approach for retrieving \emph{evidence sentences}: given a statement, (1) constructing one query per sentence and retrieving relevant articles from Wikipedia, and (2) reranking paragraphs and then sentences to create the final set of evidence sentences. 
Wikipedia is used as our evidence source mainly due to its objective perspective and broad coverage of topics. A dump of December 21, 2016 was downloaded. 
For training, evidence sentences are retrieved with queries constructed from target user arguments. For test, queries are constructed from OP.

\smallskip
\noindent \textbf{Article Retrieval.} 
We first create an inverted index lookup table for Wikipedia as done in \newcite{chen-EtAl:2017:Long4}. 
For a given statement, we construct one query per sentence to broaden the diversity of retrieved articles. Therefore, multiple passes of retrieval will be conducted if more than one query is created. 
Specifically, we first collect topic signature words of the post. Topic signatures~\cite{lin2000automated} are terms strongly correlated with a given post, measured by log-likelihood ratio against a background corpus. We treat posts from other discussions in our dataset as background. 
For each sentence, one query is constructed based on the noun phrases and verbs containing at least one topic signature word. 
For instance, a query ``\texttt{the government}, \texttt{my e-mails}, \texttt{national security}'' is constructed for the first sentence of OP in the motivating example (Figure~\ref{fig:pipeline}). 
Top five retrieved articles with highest TF-IDF similarity scores are kept per query.

\smallskip
\noindent \textbf{Sentence Reranking.} 
The retrieved articles are first segmented into paragraphs, which are reranked by TF-IDF similarity to the given statement. Up to 100 top ranked paragraphs with positive scores are retained. These paragraphs are further segmented into sentences, and reranked according to TF-IDF similarity again. We only keep up to 10 top sentences with positive scores for inclusion in the evidence set.

\begin{table}[t]
\fontsize{10}{12}\selectfont
    \centering
    \setlength{\tabcolsep}{2mm}
    \begin{tabular}{l|p{15mm} p{15mm}}
    \hline
         & \multicolumn{2}{|c}{\textit{Queries Constructed from}}\\
         & \textbf{OP} & \textbf{Argument}   \\       
        \hline
        Avg \# Topic Sig. & 17.2 & 9.8 \\
        Avg \# Query	& 6.7 &	1.9 \\
      Avg \# Article Retrieved 	  & 26.1 &	8.0 \\
       Avg \# Sent. Retrieved 	  & 67.3 &	8.5 \\
        \hline
    \end{tabular}

    \caption{\fontsize{10}{12}\selectfont 
    Statistics for evidence sentence retrieval from Wikipedia. Considering query construction from either OP or target user arguments, we show the average numbers of topic signatures collected, queries constructed, and retrieved articles and sentences. 
    }
    \label{tab:retrieval-stats}
\end{table}

\subsection{Gold-Standard Keyphrase Construction}
To create training data for the keyphrase decoder, we use the following rules to identify  keyphrases from evidence sentences that are reused by human writers for argument construction:

{
\begin{itemize}
\vspace{-2mm}
\item Extract noun phrases and verb phrases from evidence sentences using Stanford CoreNLP~\cite{manning-EtAl:2014:P14-5}.
\vspace{-2mm}
\item Keep phrases of length between 2 and 10 that overlap with content words in the argument.
\vspace{-2mm}
\item If there is span overlap between phrases, the longer one is kept if it has more content word coverage of the argument; otherwise the shorter one is retained.
\end{itemize}
}

The resultant phrases are then concatenated with a special delimiter \texttt{<phrase>} and used as gold-standard generation for training.

\section{Experimental Setup}
\label{sec:experiments}
\subsection{Final Dataset Statistics}
Encoding the full set of evidence by our current decoder takes a huge amount of time. 
We there propose a sampling strategy to allow the encoder to finish encoding within reasonable time by considering only a subset of the evidence: For each sentence in the statement, up to three evidence sentences are randomly sampled from the retrieved set; then the sampled sentences are concatenated. 
This procedure is repeated three times per statement, where a statement is an user argument for training data and an OP for test set. In our experiments, we remove duplicates samples and the ones without any retrieved evidence sentence. 
Finally, we break down the augmented data into a training set of 224,553 examples (9,737 unique OPs), 13,911 for validation (640 OPs), and 30,417 retained for test (1,892 OPs).

\subsection{Training Setup}
For all models, we use a two-layer biLSTM as encoder and a two-layer unidirectional LSTM as decoder, with 200-dimensional hidden states in each layer. We apply dropout~\cite{gal2016theoretically} on RNN cells with a keep probability of 0.8. 
We use Adam~\cite{kingma2014adam} with an initial learning rate of 0.001 to optimize the cross-entropy loss. Gradient clipping is also applied with the maximum norm of 2. 
The input and output vocabulary sizes are both 50k. 

\smallskip
\noindent \textbf{Curriculum Training.} 
We train the models in three stages where the truncated input and output lengths are gradually increased. Details are listed in Table \ref{tab:staged-training}. Importantly, this strategy allows model training to make rapid progress during early stages. 
Training each of our full models takes about 4 days on a Quadro P5000 GPU card with a batch size of 32. The model converges after about 10 epochs in total with pre-training initialization, which is described below.

\begin{table}[H]
    \centering
    \fontsize{10}{12}\selectfont
    \begin{tabular}{l|c|c|c}
    \hline
        \textbf{Component} & \textbf{Stage 1} & \textbf{Stage 2} & \textbf{Stage 3}   \\
        \hline
        \multicolumn{4}{l}{\it {Encoder}}\\
        \quad OP &  50 & 150 & 400 \\
        \quad Evidence 	&  0  &  80 & 120 \\
        \multicolumn{4}{l}{\it {Decoder}}\\
        \quad Keyphrases & 0 & 80 & 120 \\
        \quad Target Argument 	& 30 & 80 & 120 \\
    \hline
    \end{tabular}
    \caption{\fontsize{10}{12}\selectfont
    Truncation size (i.e., number of tokens including delimiters) for different stages during training. Note that in the first stage we do not include evidence and keyphrases.}
    \label{tab:staged-training}
\end{table}

\smallskip
\noindent \textbf{Adding Pre-training.} 
We pre-train a two-layer seq2seq model with OP as input and target argument as output from our training set. After 20 epochs (before converging), parameters for the first layer are used to initialize the first layer of all comparison models and our models (except for the keyphrase decoder). Experimental results show that pre-training boosts all methods by roughly 2 METEOR~\cite{denkowski-lavie:2014:W14-33} points. We describe more detailed results in the supplementary material.

\subsection{Baseline and Comparisons}

We first consider a \textsc{Retrieval}-based baseline, which concatenates retrieved evidence sentences to form the argument. 
We further compare with three seq2seq-based generation models with different training data: 
(1) \textsc{seq2seq}: training with OP as input and the argument as output; 
(2) \textsc{seq2seq} + \textit{encode evd}: augmenting input with evidence sentences as in our model; 
(3) \textsc{seq2seq} + \textit{encode KP}: augmenting input with gold-standard keyphrases, which assumes some of the talking points are known. 
All seq2seq models use a regular beam search decoder with the same beam size as ours. 

\smallskip
\noindent \textbf{Variants of Our Models.} 
We experiment with variants of our models based on the proposed separate decoder model (\textsc{Dec-separate}) or using a shared decoder (\textsc{Dec-shared}). For each, we further test whether adding keyphrase attention for argument decoding is helpful (+ \textit{attend KP}).

\smallskip
\noindent \textbf{System vs. Oracle Retrieval.} 
For test time, evidence sentences are retrieved with queries constructed from OP (\textit{System Retrieval}). We also experiment with an \textit{Oracle Retrieval} setup, where the evidence is retrieved based on user arguments, to indicate how much gain can be expected with better retrieval results.

\section{Results}
\label{sec:results}
\subsection{Automatic Evaluation}
\label{sec:autoeval}
For automatic evaluation, we use BLEU~\cite{papineni-EtAl:2002:ACL}, an $n$-gram precision-based metric (up to bigrams are considered), and METEOR~\cite{denkowski-lavie:2014:W14-33}, measuring unigram recall and precision by considering paraphrases, synonyms, and stemming. Human arguments are used as the gold-standard. 
Because each OP may be paired with more than one high-quality arguments, we compute BLEU and METEOR scores for the system argument compared against all arguments, and report the best. We do not use multiple reference evaluation because the arguments are often constructed from different angles and cover distinct aspects of the issue. 
For models that generate more than one arguments based on different sets of sampled evidence, the one with the highest score is considered. 

\begin{table}[t]
\fontsize{9}{11}\selectfont
 \setlength{\tabcolsep}{0.6mm}
   \centering
  \begin{tabular}{|l|ccc|ccc|}
  \hline
     & \multicolumn{3}{|c|}{\textit{w/ System Retrieval}} & \multicolumn{3}{|c|}{\textit{w/ Oracle Retrieval}} \\
     & \textbf{BLEU} & \textbf{MTR} & \textbf{Len}  & \textbf{BLEU} & \textbf{MTR} & \textbf{Len} \\
     \hline
  \multicolumn{7}{|l|}{\bf Baseline} \\
  \textsc{Retrieval}  & 15.32 & \textbf{12.19} & 151.2 & 10.24 & \textbf{16.22} & 132.7 \\
  \hline
  \multicolumn{7}{|l|}{\bf Comparisons} \\
  \textsc{Seq2seq} & 10.21 & 5.74 & 34.9 &7.44 & 5.25 &31.1 \\
  \quad + \textit{encode evd} & 18.03 & 7.32 & 67.0 & 13.79 & 10.06 & 68.1 \\
  \quad + \textit{encode KP} & 21.94 & 8.63 & 74.4 & 12.96 & 10.50 & 78.2 \\
  
  \hline
  \multicolumn{7}{|l|}{\bf Our Models}\\
  \textsc{Dec-shared} & 21.22 & 8.91 &  69.1 & 15.78 & 11.52 &  68.2 \\
  \quad + \textit{attend KP}  & \textbf{24.71} & 10.05 & 74.8  & 11.48 & 10.08 &  40.5 \\
  \textsc{Dec-separate} & 24.24 & 10.63 &  88.6  & 17.48 & 13.15 & 86.9 \\
  \quad + \textit{attend KP} & 24.52 & 11.27 & 88.3  & \textbf{17.80} & 13.67 & 86.8 \\
 \hline

  \end{tabular}
  \caption{\fontsize{10}{12}\selectfont  
  Results on argument generation by BLEU and METEOR (MTR), with system retrieved evidence and oracle retrieval. The best performing model is highlighted in \textbf{bold} per metric. Our separate decoder models, with and without keyphrase attention, statistically significantly outperform all seq2seq-based models based on approximation randomization testing~\cite{noreen1989computer}, $p < 0.0001$.}
  \label{tab:main-results-sampling}
\end{table}

As can be seen from Table~\ref{tab:main-results-sampling}, our models produce better BLEU scores than almost all the comparisons. Especially, our models with separate decoder yield significantly higher BLEU and METEOR scores than all seq2seq-based models (approximation randomization testing, $p < 0.0001$) do. Better METEOR scores are achieved by the \textsc{Retrieval} baseline, mainly due to its significantly longer arguments. 

Moreover, utilizing attention over both input and the generated keyphrases further boosts our models' performance. 
Interestingly, utilizing system retrieved evidence yields better BLEU scores than using oracle retrieval for testing. The reason could be that arguments generated based on system retrieval contain less topic-specific words and more generic argumentative phrases. Since the later is often observed in human written arguments, it may lead to higher precision and thus better BLEU scores.

\begin{figure}[t]
	\centering
    \includegraphics[width=75mm]{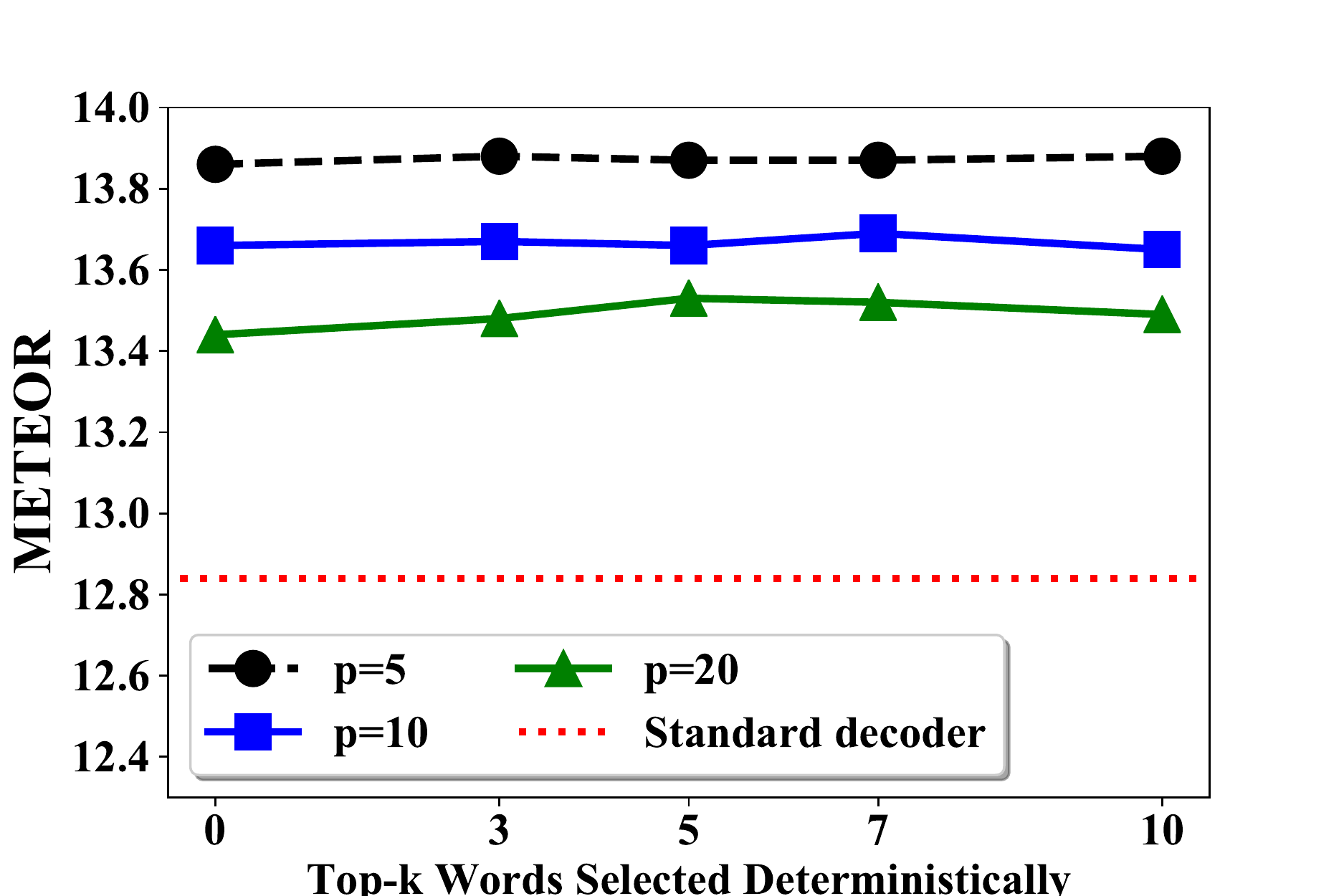}
    \caption{\fontsize{10}{12}\selectfont  
    Effect of our reranking-based decoder. Beams are reranked at every 5, 10, and 20 steps ($p$). For each step size, we also show the effect of varying $k$, where top-$k$ words are selected deterministically for beam expansion, with $10-k$ randomly sampled over multinomial distribution after removing the $k$ words. Reranking with smaller step size yields better results.}
    \label{fig:dec-comparison}
\end{figure}

\smallskip
\noindent \textbf{Decoder Strategy Comparison.} 
We also study the effect of our reranking-based decoder by varying the reranking step size ($p$) and the number of top words expanded to beam hypotheses deterministically ($k$). From the results in Figure~\ref{fig:dec-comparison}, we find that reranking with a smaller step size, e.g., $p=5$, can generally lead to better METEOR scores. Although varying the number of top words for beam expansion does not yield significant difference, we do observe more diverse beams from the system output if more candidate words are selected stochastically (i.e. with a smaller $k$).

\subsection{Topic-Relevance Evaluation}
\label{sec:topiceval}
During our pilot study, we observe that generic arguments, such as ``I don't agree with you'' or ``this is not true'', are prevalent among generations by seq2seq models. We believe that good arguments should include content that addresses the given topic. Therefore, we design a novel evaluation method to measure whether the generated arguments contain topic-relevant information.

To achieve the goal, we first train a topic-relevance estimation model inspired by the latent semantic model in~\newcite{huang2013learning}. A pair of OP and argument, each represented as the average of word embeddings, are separately fed into a two-layer transformation model. A dot-product is computed over the two projected low-dimensional vectors, and then a sigmoid function outputs the relevance score. 
For model learning, we further divide our current training data into training, developing, and test sets. 
For each OP and argument pair, we first randomly sample 100 arguments from other threads, and then pick the top 5 dissimilar ones, measured by Jaccard distance, as negative training samples. This model achieves a Mean Reciprocal Rank (MRR) score of 0.95 on the test set. Descriptions about model formulation and related training details are included in the supplementary material. 

We then take this trained model to evaluate the relevance between OP and the corresponding system arguments. Each system argument is treated as positive sample; we then select five negative samples from arguments generated for other OPs whose evidence sentences most similar to that of the positive sample. 
Intuitively, if an argument contains more topic relevant information, then the relevance estimation model will output a higher score for it; otherwise, the argument will receive a lower similarity score, and thus cannot be easily distinguished from negative samples. 
Ranking metrics of MRR and Precision at 1 (P@1) are utilized, with results reported in Table \ref{tab:ranker-result}. The ranker yields significantly better scores over arguments generated from models trained with evidence, compared to arguments generated by \textsc{seq2seq} model. 
\begin{table}[t]
    \centering
    \fontsize{9}{11}\selectfont
    \setlength{\tabcolsep}{1mm}
    \begin{tabular}{|l|c c|c c|}
    \hline
         & \multicolumn{2}{|c|}{\bf Standard Decoder} & \multicolumn{2}{|c|}{\bf Our Decoder}   \\
         & MRR & P@1 & MRR & P@1 \\
        \hline
        
        \multicolumn{5}{|l|}{\bf Baseline} \\
        \textsc{Retrieval} & 81.08 & 65.45 & - & - \\
        \hline
        \multicolumn{5}{|l|}{\bf Comparisons} \\
        \textsc{Seq2seq} & 75.29 & 58.85 & 74.46 & 57.06 \\
        \quad + \textit{encode evd} & 83.73 & 71.59 & 88.24 & 78.76 \\
        \hline
        
        \multicolumn{5}{|l|}{\bf Our Models}\\
        \textsc{Dec-shared} & 79.80 & 65.57 & \textbf{95.18} & \textbf{90.91} \\
        \quad + \textit{attend KP} & \textbf{94.33} & \textbf{89.76} & 93.48 & 87.91\\
        \textsc{Dec-separate} & 86.85 & 76.74 & 91.70 & 84.72 \\
        \quad + \textit{attend KP} & 88.53 & 79.05 & 92.77 & 86.46 \\
        \hline
    \end{tabular}
    \caption{\fontsize{10}{12}\selectfont 
    Evaluation on topic relevance---models that generate arguments highly related with OP should be ranked high by a separately trained relevance estimation model, i.e., higher Mean Reciprocal Rank (MRR) and Precision at 1 (P@1) scores. All models trained with evidence significantly outperform seq2seq trained without evidence (approximation randomization testing, $p < 0.0001$).}
    \label{tab:ranker-result}
\end{table}

Moreover, we manually pick 29 commonly used generic responses (e.g., ``I don't think so") and count their frequency in system outputs. For the seq2seq model, more than 75\% of its outputs contain at least one generic argument, compared to 16.2\% by our separate decoder model with attention over keyphrases. This further implies that our model generates more topic-relevant content.

\subsection{Human Evaluation}
\label{sec:humaneval}
We also hire three trained human judges who are fluent English speakers to rate system arguments for the following three aspects on a scale of 1 to 5 (with 5 as best): \textit{Grammaticality}---whether an argument is fluent, \textit{informativeness}---whether the argument contains useful information and is not generic, and \textit{relevance}---whether the argument contains information of a different stance or off-topic. 30 CMV threads are randomly selected, each of which is presented with randomly-shuffled OP statement and four system arguments. 

Table \ref{tab:human-evaluation} shows that our model with separate decoder and attention over keyphrases produce significantly more informative and relevant arguments than seq2seq trained without evidence.\footnote{Inter-rater agreement scores for these three aspects are 0.50, 0.60, and 0.48 by Krippendorff's $\alpha$.} 
However, we also observe that human judges prefer the retrieved arguments over generation-based models, illustrating the gap between system arguments and human edited text. 
Sample arguments are displayed in Figure~\ref{fig:sample-outputs}.

\begin{table}[t]
    \centering
    \fontsize{10}{12}\selectfont
    \begin{tabular}{|l|c|c|c|}
    \hline
        {\bf System} & {\bf Gram} & {\bf Info} & {\bf Rel}  \\
        \hline
        \textsc{Retrieval} & \textbf{4.5} $\pm$ 0.6 & \textbf{3.7} $\pm$ 0.9 & \textbf{3.3} $\pm$ 1.1 \\
        \textsc{Seq2seq} & 3.3 $\pm$ 1.1 & 1.2 $\pm$ 0.5 & 1.4 $\pm$ 0.7  \\
        \textsc{Our Model} & 2.5 $\pm$ 0.8 & 1.6 $\pm$ 0.8 & 1.8 $\pm$ 0.8   \\
        \hline
    \end{tabular}
    \caption{\fontsize{10}{12}\selectfont 
    Human evaluation results on grammaticality (\textbf{Gram}), informativeness (\textbf{Info}), and relevance (\textbf{Rel}) of arguments. 
    Our model with separate decoder and attention over keyphrases receives significantly better ratings in informativeness and relevance than seq2seq (one-way ANOVA, $p < 0.005$). 
    }
    \label{tab:human-evaluation}
\end{table}

\begin{figure}[th]
	\fontsize{8}{9}\selectfont
     \hspace{-2mm}
	\setlength{\tabcolsep}{0.8mm}
	\begin{tabular}{|p{78mm}|}
	\hline
	\textbf{Input}: 
    Putin is trying to re-form a ``Soviet Union'' with his past actions in Georgia and current actions in Ukraine \\
	\textbf{Human}: So far Putin has only really targeted areas where the local population is majority russian (or historically majority russian) and sympathetic to russia. That's not to say that he wouldn't go further, but I don't think there is yet any compelling reason to believe he would. \\
    \textbf{Retrieval}: In 1968, the Soviet Union and Warsaw pact allies invaded Czechoslovakia to halt the Prague spring reforms. It was this significant lack of flexibility that was exploited by the US to undermine the strength of the Soviet Union and thus foster its reform. In the summer of 1991, the new union treaty, which would have turned the Soviet Union into a much looser union, was agreed upon by eight republics.\\
    \textbf{Our Model}: There are two reasons you are so far beyond. There is no reason to see the military army. You can infer what they assume it, so they tend to protect up. There. If we assume it were bad, they can not be controlled. So we're talking going to ignore the wars of the world. The fact of the matter is not bad. I would also assume it. However, the government can not be reminded of the world.  \\
    \textbf{Seq2seq + evd}: 1. The reason is that the government isn't the power of the war of Russia, at the time of the war of Russia. In Russia, there is no way to continue to control the military. \\
    \textbf{Seq2seq + KP}: I don't think it's a bit. I don't think you're forgetting. I don't want to see the Syrian American dream in the US. I don't see the Syrian American dream of American citizens. I don't see the Syrian American dream of American citizens. I don't see the Syrian American military in the world. I don't think it wouldn't be true.\\
	\hline \hline
    \textbf{Input}: 
    Voting for either Hillary or Trump is morally reprehensible. \\
	\textbf{Human}: I don't think a vote for either of them means "I'm totally fine with..." as you say. Why is it not acceptable to pick the option whose platform most closely aligns to your values/beliefs/desires, while acknowledging neither candidate is ideal?  \\
    \textbf{Retrieval}: Vote trading under a democratic, majority-rule institution is sometimes considered morally reprehensible behavior.\\
    \textbf{Our Model}: He's going to happen. You're overestimating the more premise. He's already overestimating to this hypothetical party. He's already killed over the supreme court positions. Looking for Bernie. This is the same reasons: Clinton is not the supreme court presidency. You've already sent Trump's right to vote. The fact is that he's going to happen. He's already going to notice. However, it's a long way.  \\
    \textbf{Seq2seq + evd}: 1. If they don't vote the vote, they would not be able to vote for any candidate. They don't have the same effect on their political power. They are not voting for them. \\
    \textbf{Seq2seq + KP}: I agree with your view. I don't agree with you. I don't think it's easy to appeal to the Clintons. If you don't want to do this? \\
	\hline
	\end{tabular}
	\vspace{-3mm}
	\caption{\fontsize{10}{12}\selectfont 
    Sample arguments generated by human, our system, and seq2seq trained with evidence. Only the main thesis is shown for the input OP. System generations are manually detokenized and capitalized.}
\label{fig:sample-outputs}
\end{figure}

\section{Further Discussion}
\label{sec:discussion}
\noindent \textbf{Keyphrase Generation Analysis.} 
Here we provide further analysis over the generated keyphrases by our separate decoder model. First, about 10\% of the keyphrases output by our model also appear in the gold-standard (i.e., used by human arguments). 
Furthermore, 36\% of generated keyphrases are reused by our system arguments. 
With human inspection, we find that although some keyphrases are not directly reused by the argument decoder, they represent high level talking points in the argument. For instance, in the first sample argument by our model in Figure \ref{fig:sample-outputs}, keyphrases ``the motive'' and ``russian'' are generated. Although not used, they suggest the topics that the argument should stay on.

\smallskip
\noindent \textbf{Sample Arguments and Future Directions.} 
As can be seen from the sample outputs in Figure~\ref{fig:sample-outputs}, our model generally captures more relevant concepts, e.g., ``military army'' and ``wars of the world'', as discussed in the first example. Meanwhile, our model also acquires argumentative style language, though there is still a noticeable gap between system arguments and human constructed arguments. As discovered by our prior work~\citep{wangetal2017}, both topical content and language style are essential elements for high quality arguments. For future work, generation models with a better control on linguistic style need to be designed. 
As for improving coherence, we believe that discourse-aware generation models~\cite{ji-haffari-eisenstein:2016:N16-1} should also be explored in the future work to enhance text planning.

\section{Related Work}
\label{sec:related}

There is a growing interest in argumentation mining from the natural language processing research community~\cite{park-cardie:2014:W14-21,ghosh-EtAl:2014:W14-21,palau2009argumentation,niculae-park-cardie:2017:Long,eger-daxenberger-gurevych:2017:Long}. 
While argument understanding has received increasingly more attention, the area of automatic argument generation is much less studied. Early work on argument construction investigates the design of argumentation strategies~\cite{reed1996architecture,carenini-moore:2000:INLG,zukerman-mcconachy-george:2000:INLG}. 
For instance, \newcite{reed1999role} describes the first full natural language argument generation system, called Rhetorica. It however only outputs a text plan, mainly relying on heuristic rules. 
Due to the difficulty of text generation, none of the previous work represents a fully automated argument generation system. This work aims to close the gap by proposing an end-to-end trained argument construction framework.

Additionally, argument retrieval and extraction are investigated~\cite{rinott-EtAl:2015:EMNLP,hua-wang:2017:Short} to deliver relevant arguments for user-specified queries. \newcite{wachsmuth-EtAl:2017:ArgumentMining} build a search engine from arguments collected from various online debate portals. 
After the retrieval step, sentence ordering algorithms are often applied to improve coherence~\cite{sato-EtAl:2015:ACL-IJCNLP-2015-System-Demonstrations,reisert-EtAl:2015:ARG-MINING}. Nevertheless, simply merging arguments from different resources inevitably introduces redundancy. To the best of our knowledge, this is the first automatic argument generation system that can synthesize retrieved content from different articles into fluent arguments.

\section{Conclusion}
\label{sec:conclusion}
We studied the novel problem of generating arguments of a different stance for a given statement. We presented a neural argument generation framework enhanced with evidence retrieved from Wikipedia. Separate decoders were designed to first produce a set of keyphrases as talking points, and then generate the final argument. 
Both automatic evaluation against human arguments and human assessment showed that our model produced more informative arguments than popular sequence-to-sequence-based generation models.

\section*{Acknowledgements}
This work was partly supported by National Science Foundation Grant IIS-1566382, and a GPU
gift from Nvidia. 
We thank three anonymous reviewers for their insightful suggestions on various aspects of this work.

\bibliography{argument}
\bibliographystyle{acl_natbib}

\end{document}